\documentclass[journal]{IEEEtran}

\usepackage{graphicx}
\usepackage{booktabs}
\usepackage{amsmath}
\usepackage[hidelinks]{hyperref}

\graphicspath{{figs/}}

\newcommand{\ours}{StripPool-UNet+FPN}

\begin{document}

\title{Consensus-gated Multi-Agent Neural Architecture Search for Seismic Fault Segmentation}

\author{Shehram~Baig and Ahmad~Mustafa, \IEEEmembership{Member,~IEEE} %
\thanks{Shehram Baig is with the Information Technology University (ITU), Lahore, Pakistan (e-mail: shehram.baig@yahoo.com).}%
\thanks{Ahmad Mustafa is with the King Fahd University of Petroleum and Minerals, Dhahran, Saudi Arabia (e-mail: ahmad.mustafa@kfupm.edu.pk).}}

\markboth{IEEE Geoscience and Remote Sensing Letters}%
{Baig \MakeLowercase{\textit{and}} Mustafa: Multi-LLM Debate-Driven NAS for Compact Seismic Fault Segmentation}

\maketitle

\begin{abstract}
Neural networks for seismic fault segmentation are often borrowed from computer vision and medical imaging domains where they train under relatively much larger labeled data resources. Optimizing their architecture under tight labeled data budgets as are common in geophysical applications is not a trivial problem. Manually designing data-optimal architectures is time-consuming while classical neural architecture search (NAS) is restricted to
hand-crafted search spaces and large compute budgets. We present an agentic
NAS system in which a panel of three large language models (Claude, GPT-5.1,
and Gemini~2.5~Pro) debates each candidate architecture to unanimous
consensus, authors the complete PyTorch implementation, cross-reviews it, and
submits it to an automated validate-train-score loop with a hard 450K
parameter budget, keep-or-revert lineage, and a memory of failed mechanisms.
Operating on source code rather than a predefined operation menu, the search
ran on a single consumer GPU and trained only eight candidates. It discovered
\ours{}: a 425K-parameter encoder-decoder with a strip-pooling bottleneck,
squeeze-and-excitation gating, an asymmetric one-conv decoder, and a
feature-pyramid fusion neck. Trained under a protocol identical to all
baselines on sections derived from the Thebe fault dataset, it attains the
highest F1 (0.578) and IoU of all models tested while being the smallest,
outperforming a published-capacity U-Net (31M parameters, F1 0.484),
DeepLabV3-ResNet50 (39.6M, 0.516), an Attention U-Net(1.83M, 0.502). The search cost 101 LLM calls ($\sim$1.15M input / 0.39M output
tokens) and roughly one GPU-day, making consensus-gated LLM panels a
practical, low-cost route to domain-specific architecture
discovery.
\end{abstract}


\IEEEpeerreviewmaketitle

\section{Introduction}
\IEEEPARstart{F}{ault} interpretation is central to seismic structural
analysis: faults control trap geometry, fluid migration, and drilling
hazards, yet manual picking in 3-D volumes is slow and subjective. Deep
segmentation networks now dominate automatic fault detection
\cite{wu2019faultseg,an2021deep,gao2022nested,10449677}, but the architectures in use
are hand-designed, typically adapted from natural-image or biomedical
segmentation \cite{ronneberger2015unet,chen2017deeplabv3}. Whether these
designs actually suit seismic sections is rarely asked: the class imbalance
is extreme ($\sim$5.6\% fault pixels), the targets are thin oblique
structures crossing the whole section, and the large section sizes force
training at batch size one.

Neural architecture search (NAS) automates this question in principle, but
classical NAS is expensive and constrained: reinforcement-learning and
evolutionary controllers sample thousands of models
\cite{zoph2017nas,real2019regularized}, and differentiable methods relax a
fixed, hand-drawn operation menu \cite{liu2019darts}; the search space itself
remains a human design decision. A recent line of work instead lets a large
language model (LLM) edit architecture source code directly
\cite{chen2023evoprompting,nasir2024llmatic}, following the broader ``LLM as
search operator'' paradigm
\cite{yang2024opro,romera2024funsearch,ma2024eureka,huang2024mlagentbench}.
LLMs bring two assets no classical controller has: prior knowledge of
architecture research, and the ability to justify a design before it is
trained.

We report a complete, end-to-end instance of this paradigm for seismic
fault segmentation, with two systems-level contributions. First,
because a single LLM proposer can fixate on pet mechanisms, we replace it
with a three-model debate panel (Claude, GPT-5.1, Gemini~2.5~Pro) that must
reach unanimous consensus on a design specification before any code is
written; the winning model then authors the full \texttt{network.py} and the
other two review it. Training, the expensive step, is only ever spent on
designs that survived adversarial scrutiny. Second, the search loop is
engineered for honesty and efficiency: a hard 450K parameter budget enforced
before training, keep-or-revert selection on the mean of three independent
trials, a hard ban list of failed mechanisms, and a stagnation-based stopping
rule.

Our contributions are: (1)~an agentic NAS loop in which three heterogeneous
LLMs debate to unanimous consensus, author, and cross-review complete PyTorch
architectures, with automatic validation, budget enforcement, keep-or-revert
lineage, and no predefined search space; (2)~\ours{}, a discovered
425{,}177-parameter architecture combining a strip-pooling bottleneck
\cite{hou2020strip}, squeeze-and-excitation gating \cite{hu2018senet}, an
asymmetric single-conv decoder, and an FPN-style fusion neck \cite{lin2017fpn};
(3)~an identical-protocol benchmark on Thebe-derived sections
\cite{an2021gigabyte} where the discovered network is simultaneously the
smallest and most accurate model tested, beating the published-capacity U-Net
\cite{ronneberger2015unet} (73$\times$ more parameters), DeepLabV3-ResNet50
\cite{chen2017deeplabv3,he2016resnet} (93$\times$), and Attention UNet \cite{oktay2018attention} (4$\times$); and (4)~a full cost accounting (101 LLM
calls, 1.15M input / 0.39M output tokens, $\approx$one GPU-day), plus two
cautionary findings: optimizing recall alone silently discards the best F1
designs, and published-capacity models underperform compact ones in this
regime.

\section{Related Work}
\subsection{Classical NAS}
Zoph and Le \cite{zoph2017nas} trained an RNN
controller with reinforcement learning over thousands of sampled
architectures; regularized evolution \cite{real2019regularized} achieved
similar results with aging tournaments. DARTS \cite{liu2019darts} made the
search differentiable but only within a fixed cell-level operation menu. All
three inherit a human-designed encoding; none can invent a mechanism outside
it.

\subsection{LLM-driven search} 
EvoPrompting \cite{chen2023evoprompting} used an
LLM as the mutation/crossover operator over architecture source code;
LLMatic \cite{nasir2024llmatic} paired LLM edits with a quality-diversity
archive. Beyond NAS, OPRO \cite{yang2024opro} showed LLMs act as generic
history-conditioned optimizers, FunSearch \cite{romera2024funsearch} and
Eureka \cite{ma2024eureka} evolved programs and reward functions, and
MLAgentBench \cite{huang2024mlagentbench} benchmarked autonomous ML
experimentation. Our system differs in three ways: the proposer is a
heterogeneous multi-model panel gated by unanimous consensus rather than a
single model; each accepted design is a single reasoned mutation on a
maintained lineage with explicit reverts; and the target task is dense
prediction on real geophysical data rather than classification benchmarks.
\subsection{Deep learning for seismic faults} 
FaultSeg3D \cite{wu2019faultseg}
trained a compact 3-D U-Net on synthetic volumes and remains the canonical
baseline. An \emph{et al.} released the Thebe dataset, the largest
expert-interpreted open fault benchmark, and trained U-Net and DeepLab-family
models on it \cite{an2021gigabyte,an2021deep}; a recent large-scale study
benchmarked many segmentation families on it \cite{quesada2025benchmark}.
Nested residual U-Nets \cite{gao2022nested} and facies benchmarks
\cite{alaudah2019facies} round out the domain. To our knowledge no prior work
has applied architecture search, LLM-driven or otherwise, to this task.

\section{Method}

\subsection{Task and Data}
We use two disjoint volumes (train and test) cropped from the Thebe fault
dataset \cite{an2021gigabyte} (Exmouth Plateau, NW~Australia), each
containing 100 vertical 2-D sections of $600 \times 1500$
samples with expert binary fault masks; $\sim$5.6\% of pixels are fault.
Amplitudes are min-max scaled to $[0,1]$ per volume. The large sections force
batch-size-1 training, which penalizes BatchNorm-based designs and motivates
GroupNorm \cite{wu2018groupnorm} throughout. No augmentation is used,
keeping the comparison purely architectural.

\subsection{Search Loop}
The orchestrator maintains a single lineage: run~0 trains the
baseline (1.49M parameters, budget-exempt as a reference), and each subsequent run asks the panel for a
single architectural mutation of the current parent. Every candidate passes a
validator before any GPU time is spent: the code is AST-parsed and
instantiated, must run a train-mode forward/backward pass at batch size 1 on
three section shapes ($256{\times}256$, $400{\times}800$,
$600{\times}1504$), preserve spatial dimensions, and contain at most
450{,}000 parameters. Validated candidates are trained from scratch and
scored (Sec.~\ref{sec:protocol}); a mutation is kept only if its mean
F1 over three independent trials beats the parent's, otherwise it is reverted and its core mechanism enters a hard ban list the panel may
not reintroduce. Parent selection always uses mean-of-trials (never
best-of-trials) to avoid promoting lucky seeds. The search stops at a hard
budget of 15 mutation runs or on a stagnation rule: once the last three
completed runs all beat the baseline mean yet the running-best mean F1 has
improved by less than 0.01 across that window; no target score is used. The
stagnation rule terminated the search here, after run~9.

Each run's feedback to the panel comprises the full JSON history (every prior
mutation, its rationale, per-trial scores, parameter count, and verdict), the
complete source of the current parent, the ban list, lessons imported from
archived earlier rounds, and the published-baseline scoreboard with the
two-axis objective: fewer parameters than every baseline and higher F1.

\begin{figure}[t]
\centering
\includegraphics[width=0.6\columnwidth]{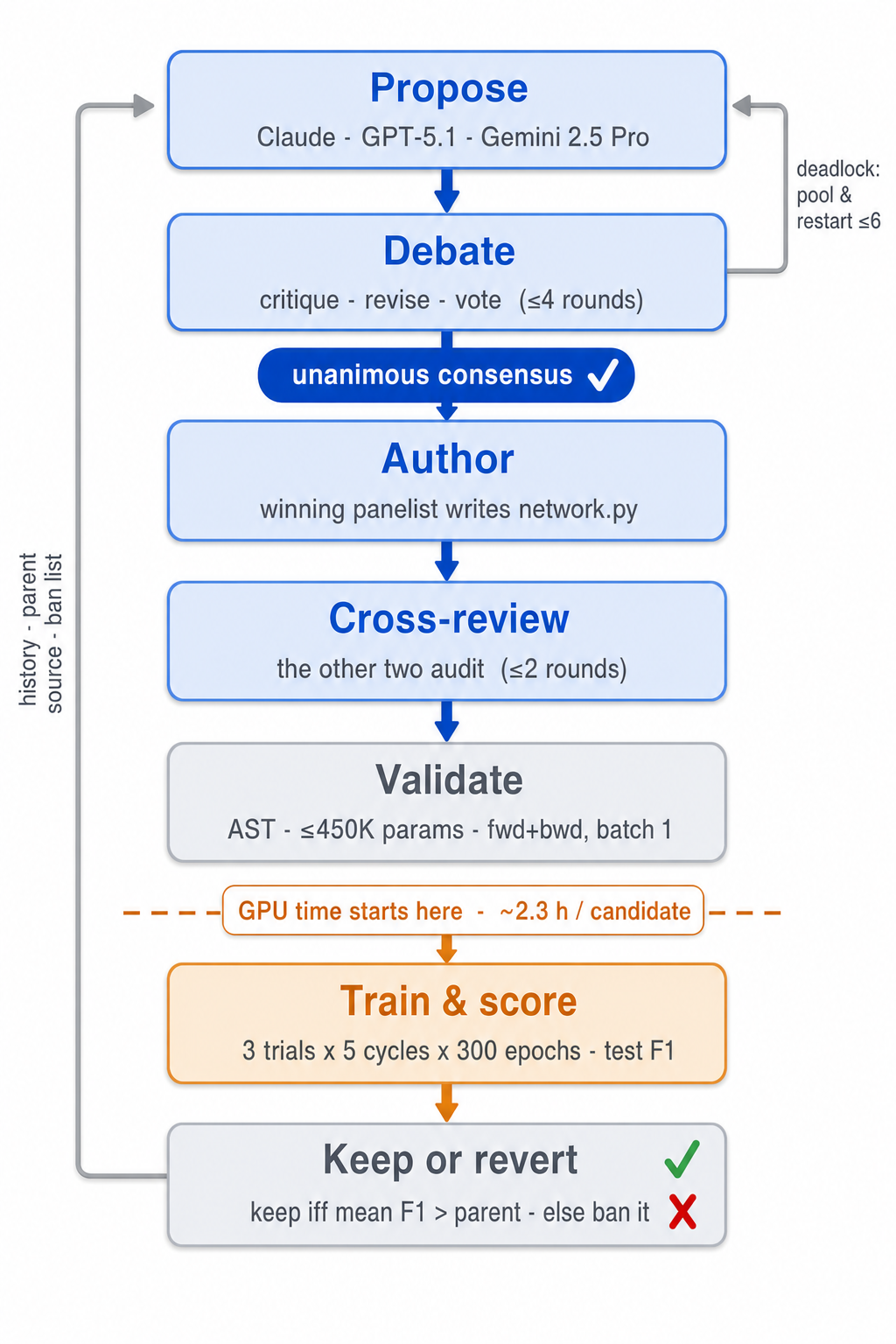}
\caption{One search iteration. Training, the expensive step, is spent only on
designs that survived unanimous three-model consensus, cross-review, and the
pre-training validator.}
\label{fig:pipeline}
\end{figure}

\subsection{The Debate Panel}
Each mutation is designed by three frontier models from different providers
(Anthropic's Claude, OpenAI's GPT-5.1, Google's Gemini~2.5~Pro) following
Fig.~\ref{fig:pipeline}. In the
\textsc{propose} phase each model independently emits a structured
specification: design name, an explicit \textsc{add/remove/change} delta
versus the parent, topology, key mechanisms, expected parameter count,
differences from banned failures, and risks. In the \textsc{debate} phase (up
to four rounds) each model critiques the other two by name, may adopt a
rival's design, and casts two votes: \textsc{endorse} (which spec to build)
and \textsc{approve} (boolean). Consensus requires all three to approve and
endorse the same specification. On deadlock, the disagreements are pooled
into a shared problem statement and the debate restarts from fresh proposals
(up to six restarts); if consensus is never reached the run is abandoned at
zero GPU cost. The endorsed model then writes the complete implementation,
and the other two review it against the training contract (interface, spatial
contract, batch-1 hazards, timeout risks) for up to two revision rounds.
Every proposal, critique, vote, and review is archived, and per-provider
token usage is logged.

\subsection{Training Protocol and Metrics}
\label{sec:protocol}
All training, for search candidates and every baseline alike, uses the same
fixed recipe: Adam \cite{kingma2015adam} with learning rate $10^{-3}$, batch
size 1, and the loss
$\mathcal{L} = \mathrm{CE}_{w}(\hat{y}, y) + \mathrm{MSE}(\hat{x}, x)$, a
class-weighted cross-entropy (fault weight 15) plus an auxiliary
image-reconstruction term, a pairing proven effective in prior seismic
interpretation work \cite{mustafa2023active,wu2023mtlfaultnet}; every network
therefore exposes a segmentation head and a one-channel reconstruction head.
No early stopping, scheduling, or augmentation is used. Training ran on the
GPU of a single Apple M-series laptop (MPS backend, FP32).

During the search each candidate trains on only five
fixed, evenly spaced sections of the training volume and is evaluated on the
full test volume, for 3 trials $\times$ 5 cycles $\times$ 300 epochs; each
cycle is an independent training from fresh random weights. Batch-1 training
on five sections is too noisy for single runs to separate architectures, so
the cycles act as random restarts: a trial scores the best of its five
restarts, estimating what the architecture achieves when optimization
succeeds, while the keep/revert decision uses the mean of the three trial
scores, which no single lucky seed can inflate. One candidate costs
$\approx$2.3 GPU-hours.

Final comparisons train every architecture on
all 100 training sections for 60 epochs under the identical loss and
optimizer, and evaluate on the full test volume.

The headline metric is pixel-wise F1 (equivalently Dice) at threshold 0.5;
we also report IoU, precision, recall, and PR-AUC (average precision).
Because $\sim$94\% of pixels are background, accuracy and recall alone are
uninformative.

\section{Results}
\subsection{Discovered Architecture}
\begin{figure*}[t]
\centering
\includegraphics[width=0.8\textwidth]{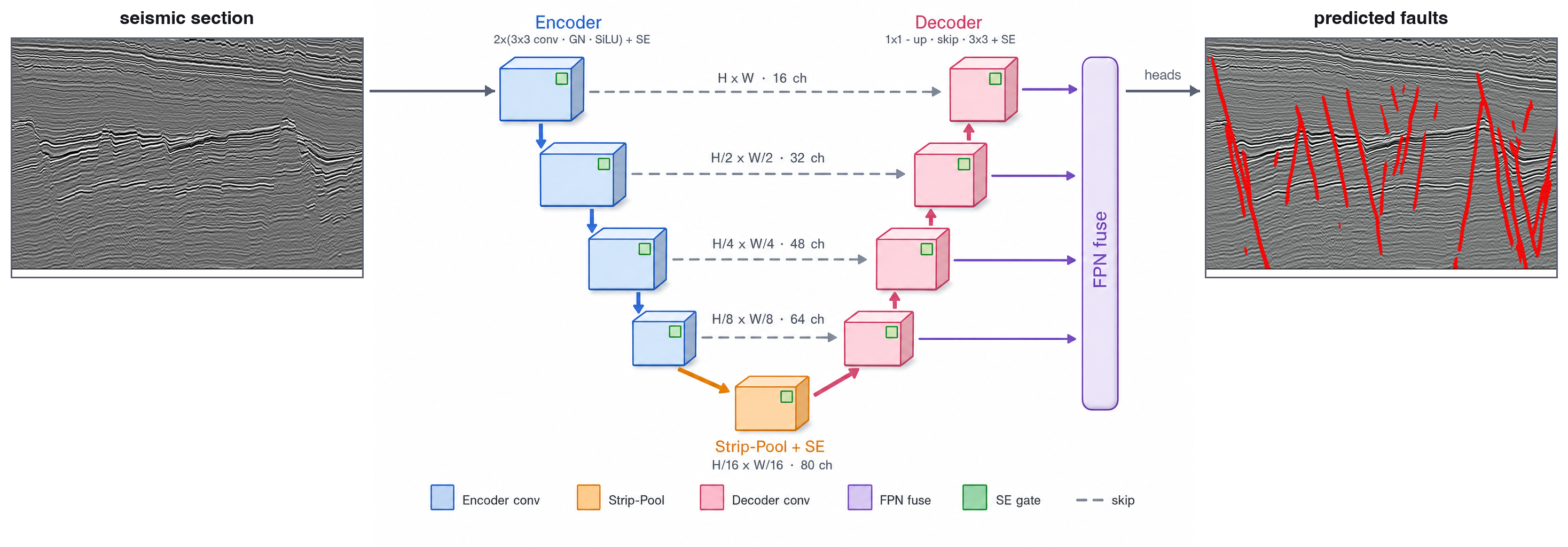}
\caption{The discovered \ours{} architecture (425{,}177 parameters): a narrow
SE-gated encoder (16/32/48/64/80 channels), a strip-pooling bottleneck at
$H/16$, an asymmetric single-conv decoder, and an FPN neck that fuses the
four decoder scales into the map read by both output heads (Sec.~IV). The
output panel overlays predicted faults (red, threshold 0.5) on a test
section.}
\label{fig:arch}
\end{figure*}

\begin{figure}[t]
\centering
\includegraphics[width=0.85\columnwidth]{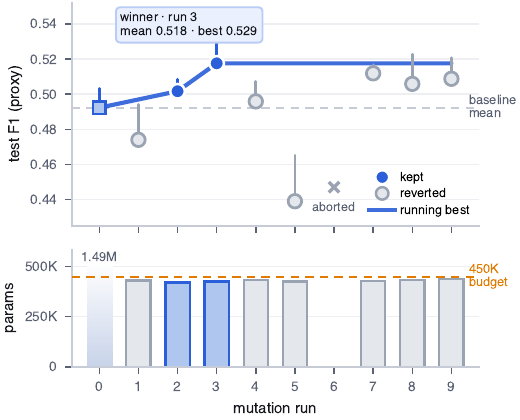}
\caption{Round-4 search trajectory. Top: test F1 under the 5-section proxy;
each marker is the mean over 3 trials, the whisker rises to the best single
trial; a long whisker flags a seed-sensitive design (e.g.\ run~8). Filled
markers were kept, open markers
reverted; the gate compares means against the current parent (the
running-best line), not the baseline, which is why runs 7--9 clear the
dashed baseline yet were rejected. Run 6 aborted during the debate on a
provider transport failure. Bottom: parameter count against the 450K budget;
the 1.49M reference baseline is exempt.}
\label{fig:traj}
\end{figure}

The winning design, \ours{} (Fig.~\ref{fig:arch}; run~3 of
Fig.~\ref{fig:traj}; 425{,}177 parameters), is a U-Net-shaped encoder-decoder
with four asymmetries the panel argued for explicitly:

Narrow full-rank encoder. Five stages at 16/32/48/64/80 channels,
each two $3{\times}3$ convolutions with GroupNorm \cite{wu2018groupnorm} and
SiLU, followed by squeeze-and-excitation (SE) gating \cite{hu2018senet}
(reduction 8). The panel rejected depthwise-separable convolutions, the
``obvious'' compression move, after run~1 showed they regress on this task;
narrow full-rank filters spend the budget better here.

At $H/16$ the bottleneck applies a
strip-pooling module \cite{hou2020strip}: features are averaged along entire
rows and columns, mixed by $(3,1)$ and $(1,3)$ convolutions, broadcast-summed,
and turned into a sigmoid gate. This captures section-wide lateral context
(matched to faults: thin quasi-linear structures crossing the whole section)
at negligible parameter cost, where a full ASPP would not fit the budget.

Decoder blocks use a $1{\times}1$ reduction,
bilinear upsampling, skip concatenation, and a single $3{\times}3$
convolution with SE, half the encoder's depth. The panel's rationale:
representation is learned in the encoder; the decoder only needs to
relocalize it.

The four decoder scales are projected to 16 channels
by $1{\times}1$ convolutions, upsampled to full resolution, summed, and
refined by one $3{\times}3$ convolution \cite{lin2017fpn}; both output heads
read from the fused map, giving every decoder stage a direct gradient path.

\subsection{Search Behavior}
\label{sec:search}
Fig.~\ref{fig:traj} and Table~\ref{tab:runs} summarize the final search
round. Eight candidates were trained (one debate aborted on a provider
transport failure); all eight arrived under budget, and every completed
debate reached consensus in one or two rounds. Two mutations were
kept, the strip-pooling redesign (run~2) and the FPN fusion neck (run~3), and
six were reverted, including plausible ideas (coordinate priors, non-local
attention, large-kernel depthwise convolutions) that the keep-or-revert gate
caught as regressions. The search stopped on the stagnation rule after runs
7--9 failed to improve on run~3.

The winner's evidence is not a lucky seed: run~3's three per-trial F1 scores
(0.5154/0.5294/0.5081) do not overlap the 1.49M baseline's
(0.4923/0.4814/0.5033); its worst trial beats the baseline's best, at
$3.5\times$ fewer parameters.

Earlier search rounds optimized test recall: one such search improved mean
recall $3.1\times$ (0.194 $\to$ 0.594) with non-overlapping trial
distributions, but re-scoring every trained model with the full metric suite
revealed it had kept a worse model and reverted its genuinely best design (an
FPN fusion variant): with a
15$\times$ class weight and $\sim$5.6\% positives, recall rewards
over-prediction while precision collapses. All results here therefore use F1
as the search objective.

\begin{table}[t]
\caption{Final search round: one debated mutation per run
(proxy protocol, mean/best F1 over 3 trials).}
\label{tab:runs}
\centering
\scriptsize
\setlength{\tabcolsep}{2.6pt}
\begin{tabular}{@{}clrccc@{}}
\toprule
run & mutation & params & mean F1 & best F1 & verdict \\
\midrule
0 & baseline (reference) & 1{,}492{,}763 & 0.4923 & 0.5033 & ref. \\
1 & depthwise-sep.\ + ASPP-lite & 430{,}119 & 0.4740 & 0.4941 & revert \\
2 & strip-pooling U-Net & 420{,}217 & 0.5017 & 0.5085 & \textbf{keep} \\
3 & + FPN fusion neck & 425{,}177 & \textbf{0.5176} & \textbf{0.5294} & \textbf{keep} \\
4 & + per-scale strip gates & 434{,}873 & 0.4959 & 0.5073 & revert \\
5 & + coordinate priors & 425{,}753 & 0.4390 & 0.4652 & revert \\
6 & (debate aborted) & n/a & n/a & n/a & n/a \\
7 & + raw-input pyramid & 427{,}193 & 0.5119 & 0.5168 & revert \\
8 & + low-rank non-local attn. & 434{,}290 & 0.5059 & 0.5228 & revert \\
9 & + depthwise $7{\times}7$ encoder & 437{,}417 & 0.5088 & 0.5208 & revert \\
\bottomrule
\end{tabular}
\end{table}

\subsection{Identical-Protocol Benchmark}
Table~\ref{tab:bench} is the main result. Under a training protocol identical
for every row (full training volume, 60 epochs, same loss and optimizer,
GroupNorm batch-1 adaptation for all baselines), the discovered network is
simultaneously the smallest and the most accurate model: F1 0.578 and IoU
0.406 at 0.43M parameters. It outperforms the published-capacity U-Net
\cite{ronneberger2015unet} by $+0.094$ F1 at $73\times$ fewer parameters, and
DeepLabV3-ResNet50 \cite{chen2017deeplabv3,he2016resnet}, a 39.6M-parameter
model that needed $9\times$ the training wall-clock, by $+0.061$ F1 at
$93\times$ fewer. U-Net and DeepLab-family models have published Thebe
results \cite{an2021deep,quesada2025benchmark}, but those use tolerance-based
OIS/ODS edge matching on different splits and are not directly comparable,
which is precisely why we retrain every architecture under one protocol.

\subsection{Cost}
The final search consumed 101 LLM calls totaling 1{,}148{,}853 input and
388{,}377 output tokens (Claude 519K/151K, Gemini 330K/180K, GPT 300K/57K), a
few tens of US dollars at current API prices, and roughly one GPU-day on a
single laptop for the eight trained candidates. Classical NAS trains hundreds
to thousands of candidates for such results
\cite{zoph2017nas,real2019regularized}; here the debate gate meant every
trained candidate was budget-compliant and mechanically valid, and one in four improved the lineage.

\begin{table}[t]
\caption{Full-data benchmark: every model trained under the identical
protocol on Thebe-derived sections; test-volume metrics, F1-sorted.}
\label{tab:bench}
\centering
\scriptsize
\setlength{\tabcolsep}{2.0pt}
\begin{tabular}{@{}lrccccc@{}}
\toprule
model & params & F1 & IoU & Prec. & Rec. & PR-AUC \\
\midrule
\textbf{\ours{} (ours)} & \textbf{0.43M} & \textbf{0.578} & \textbf{0.406} & 0.485 & 0.715 & \textbf{0.863} \\
DeepLabV3-R50 \cite{chen2017deeplabv3} & 39.63M & 0.516 & 0.348 & 0.450 & 0.607 & 0.824 \\
Attention U-Net \cite{oktay2018attention} & 1.83M & 0.502 & 0.336 & 0.371 & 0.777 & 0.852 \\
U-Net (1.8M) \cite{ronneberger2015unet} & 1.81M & 0.494 & 0.328 & 0.372 & 0.737 & 0.829 \\
U-Net (published) \cite{ronneberger2015unet} & 31.04M & 0.484 & 0.319 & 0.381 & 0.663 & 0.813 \\
\bottomrule
\end{tabular}
\end{table}

\begin{figure*}[t]
\centering
\includegraphics[width=0.88\textwidth]{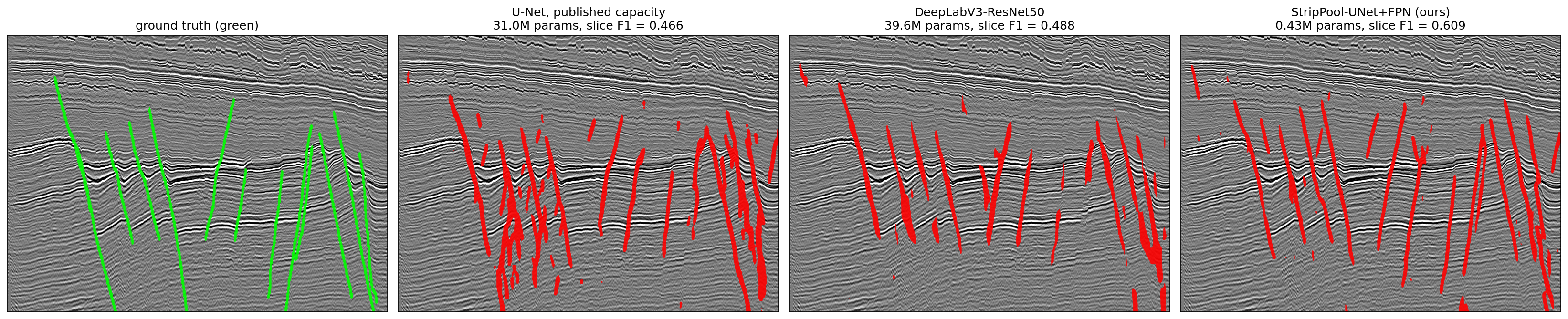}
\caption{Test-set fault predictions (threshold 0.5, identical protocol) on
the test section where the discovered network's advantage over the published
baselines is largest (per-slice F1 in each panel). Left to right: ground
truth; published-capacity U-Net (31.0M, F1 0.466); DeepLabV3-ResNet50 (39.6M,
0.488); discovered \ours{} (0.43M, 0.609). The large models draw thick,
blobby responses or fragmented strands; the compact network draws thin,
continuous, well-separated strands tracking the ground-truth geometry, and
attains the best per-slice F1 on 96 of the 100 test sections.}
\label{fig:qual}
\end{figure*}

\section{Discussion}
Table \ref{tab:bench} shows the searched model overwhelmingly outperformed the published SOTA models based on the popular UNet and DeepLab architectures trained and tested under identical settings. This is confirmed through a visual inspection of the fault predictions generated on a test set slice in Fig. \ref{fig:qual}. The searched model can be seen to produce seismic fault predictions that are cleaner and closer to the ground-truth compared to the baseline models. 

This finding highlights the potential of agentic workflows in optimizing model architecture design with respect to the constraints of the target application (in this case, the extreme class imbalance and label scarcity of the problem). Off-the-shelf segmentation models in computer vision and medical imaging domains are designed to expect hundreds to thousands of labeled training examples. Such over-parameterized models carry a high risk of under-performing when deployed in scarce-labeled data applications such as geological fault segmentation. Simply scaling down the capacity of the baseline model in a naive manner (such as in row 4 of Table \ref{tab:bench}) still can not outperform a model optimized to the target application via our proposed agentic workflow.    

A potential future direction to explore in this line of work would be to have an agent specifically judge the visual quality of the results to give feedback to the proposers in the next round. This is because quantitative metrics are often insufficient to fully reflect model performance for fault segmentation given what is often times an incomplete and uncertain nature of ground-truth labels. This would help mitigate the issue of the agentic pipeline discarding otherwise perfectly valid architectures simply based on minor regressions in quantitative performance.  

\section{Conclusion}
A consensus-gated panel of three frontier LLMs, embedded in a validated
keep-or-revert search loop with a hard parameter budget, discovered a
425K-parameter fault segmentation network that beats every published
architecture we retrained under an identical protocol, including models
73--93$\times$ its size, for about a GPU-day and a few dollars of API calls.
For specialized dense-prediction tasks, reasoned code-level architecture
search is now cheaper than scaling, and multi-model debate is an effective
quality gate for scarce training compute. 

\bibliographystyle{IEEEtran}
\bibliography{refs}

@inproceedings{ronneberger2015unet,
  author    = {Ronneberger, Olaf and Fischer, Philipp and Brox, Thomas},
  title     = {U-Net: Convolutional Networks for Biomedical Image Segmentation},
  booktitle = {Medical Image Computing and Computer-Assisted Intervention (MICCAI)},
  year      = {2015},
  pages     = {234--241}
}

@article{chen2017deeplabv3,
  author  = {Chen, Liang-Chieh and Papandreou, George and Schroff, Florian and Adam, Hartwig},
  title   = {Rethinking Atrous Convolution for Semantic Image Segmentation},
  journal = {arXiv preprint arXiv:1706.05587},
  year    = {2017}
}

@inproceedings{he2016resnet,
  author    = {He, Kaiming and Zhang, Xiangyu and Ren, Shaoqing and Sun, Jian},
  title     = {Deep Residual Learning for Image Recognition},
  booktitle = {IEEE Conference on Computer Vision and Pattern Recognition (CVPR)},
  year      = {2016},
  pages     = {770--778}
}

@article{oktay2018attention,
  author  = {Oktay, Ozan and Schlemper, Jo and Le Folgoc, Loic and Lee, Matthew and Heinrich, Mattias and Misawa, Kazunari and Mori, Kensaku and McDonagh, Steven and Hammerla, Nils Y. and Kainz, Bernhard and Glocker, Ben and Rueckert, Daniel},
  title   = {Attention {U-Net}: Learning Where to Look for the Pancreas},
  journal = {arXiv preprint arXiv:1804.03999},
  year    = {2018}
}

@article{wu2019faultseg,
  author  = {Wu, Xinming and Liang, Luming and Shi, Yunzhi and Fomel, Sergey},
  title   = {{FaultSeg3D}: Using Synthetic Data Sets to Train an End-to-End Convolutional Neural Network for 3{D} Seismic Fault Segmentation},
  journal = {Geophysics},
  volume  = {84},
  number  = {3},
  pages   = {IM35--IM45},
  year    = {2019}
}

@article{an2021gigabyte,
  author  = {An, Yu and Guo, Jiulin and Ye, Qing and Childs, Conrad and Walsh, John and Dong, Ruihuan},
  title   = {A Gigabyte Interpreted Seismic Dataset for Automatic Fault Recognition},
  journal = {Data in Brief},
  volume  = {37},
  pages   = {107219},
  year    = {2021}
}

@article{an2021deep,
  author  = {An, Yu and Guo, Jiulin and Ye, Qing and Childs, Conrad and Walsh, John and Dong, Ruihuan},
  title   = {Deep Convolutional Neural Network for Automatic Fault Recognition from 3{D} Seismic Datasets},
  journal = {Computers \& Geosciences},
  volume  = {153},
  pages   = {104776},
  year    = {2021}
}

@article{quesada2025benchmark,
  author  = {Quesada, Jorge and others},
  title   = {A Large-Scale Benchmark on Geological Fault Delineation Models},
  journal = {arXiv preprint arXiv:2505.08585},
  year    = {2025}
}

@inproceedings{zoph2017nas,
  author    = {Zoph, Barret and Le, Quoc V.},
  title     = {Neural Architecture Search with Reinforcement Learning},
  booktitle = {International Conference on Learning Representations (ICLR)},
  year      = {2017}
}

@inproceedings{real2019regularized,
  author    = {Real, Esteban and Aggarwal, Alok and Huang, Yanping and Le, Quoc V.},
  title     = {Regularized Evolution for Image Classifier Architecture Search},
  booktitle = {AAAI Conference on Artificial Intelligence},
  year      = {2019},
  pages     = {4780--4789}
}

@inproceedings{liu2019darts,
  author    = {Liu, Hanxiao and Simonyan, Karen and Yang, Yiming},
  title     = {{DARTS}: Differentiable Architecture Search},
  booktitle = {International Conference on Learning Representations (ICLR)},
  year      = {2019}
}

@inproceedings{chen2023evoprompting,
  author    = {Chen, Angelica and Dohan, David and So, David},
  title     = {{EvoPrompting}: Language Models for Code-Level Neural Architecture Search},
  booktitle = {Advances in Neural Information Processing Systems (NeurIPS)},
  year      = {2023}
}

@inproceedings{nasir2024llmatic,
  author    = {Nasir, Muhammad Umair and Earle, Sam and Togelius, Julian and James, Steven and Cleghorn, Christopher},
  title     = {{LLMatic}: Neural Architecture Search via Large Language Models and Quality Diversity Optimization},
  booktitle = {Genetic and Evolutionary Computation Conference (GECCO)},
  year      = {2024}
}

@inproceedings{yang2024opro,
  author    = {Yang, Chengrun and Wang, Xuezhi and Lu, Yifeng and Liu, Hanxiao and Le, Quoc V. and Zhou, Denny and Chen, Xinyun},
  title     = {Large Language Models as Optimizers},
  booktitle = {International Conference on Learning Representations (ICLR)},
  year      = {2024}
}

@article{romera2024funsearch,
  author  = {Romera-Paredes, Bernardino and Barekatain, Mohammadamin and Novikov, Alexander and Balog, Matej and Kumar, M. Pawan and Dupont, Emilien and Ruiz, Francisco J. R. and Ellenberg, Jordan S. and Wang, Pengming and Fawzi, Omar and Kohli, Pushmeet and Fawzi, Alhussein},
  title   = {Mathematical Discoveries from Program Search with Large Language Models},
  journal = {Nature},
  volume  = {625},
  pages   = {468--475},
  year    = {2024}
}

@inproceedings{ma2024eureka,
  author    = {Ma, Yecheng Jason and Liang, William and Wang, Guanzhi and Huang, De-An and Bastani, Osbert and Jayaraman, Dinesh and Zhu, Yuke and Fan, Linxi and Anandkumar, Anima},
  title     = {Eureka: Human-Level Reward Design via Coding Large Language Models},
  booktitle = {International Conference on Learning Representations (ICLR)},
  year      = {2024}
}

@inproceedings{huang2024mlagentbench,
  author    = {Huang, Qian and Vora, Jian and Liang, Percy and Leskovec, Jure},
  title     = {{MLAgentBench}: Evaluating Language Agents on Machine Learning Experimentation},
  booktitle = {International Conference on Machine Learning (ICML)},
  year      = {2024}
}

@inproceedings{hu2018senet,
  author    = {Hu, Jie and Shen, Li and Sun, Gang},
  title     = {Squeeze-and-Excitation Networks},
  booktitle = {IEEE Conference on Computer Vision and Pattern Recognition (CVPR)},
  year      = {2018},
  pages     = {7132--7141}
}

@inproceedings{wu2018groupnorm,
  author    = {Wu, Yuxin and He, Kaiming},
  title     = {Group Normalization},
  booktitle = {European Conference on Computer Vision (ECCV)},
  year      = {2018},
  pages     = {3--19}
}

@inproceedings{lin2017fpn,
  author    = {Lin, Tsung-Yi and Doll{\'a}r, Piotr and Girshick, Ross and He, Kaiming and Hariharan, Bharath and Belongie, Serge},
  title     = {Feature Pyramid Networks for Object Detection},
  booktitle = {IEEE Conference on Computer Vision and Pattern Recognition (CVPR)},
  year      = {2017},
  pages     = {2117--2125}
}

@inproceedings{hou2020strip,
  author    = {Hou, Qibin and Zhang, Li and Cheng, Ming-Ming and Feng, Jiashi},
  title     = {Strip Pooling: Rethinking Spatial Pooling for Scene Parsing},
  booktitle = {IEEE/CVF Conference on Computer Vision and Pattern Recognition (CVPR)},
  year      = {2020},
  pages     = {4003--4012}
}

@article{alaudah2019facies,
  author  = {Alaudah, Yazeed and Micha{\l}owicz, Patrycja and Alfarraj, Motaz and AlRegib, Ghassan},
  title   = {A Machine-Learning Benchmark for Facies Classification},
  journal = {Interpretation},
  volume  = {7},
  number  = {3},
  pages   = {SE175--SE187},
  year    = {2019}
}

@article{gao2022nested,
  author  = {Gao, Kai and Huang, Lianjie and Zheng, Yingcai},
  title   = {Fault Detection on Seismic Structural Images Using a Nested Residual {U-Net}},
  journal = {IEEE Transactions on Geoscience and Remote Sensing},
  volume  = {60},
  pages   = {1--15},
  year    = {2022}
}

@inproceedings{kingma2015adam,
  author    = {Kingma, Diederik P. and Ba, Jimmy},
  title     = {Adam: A Method for Stochastic Optimization},
  booktitle = {International Conference on Learning Representations (ICLR)},
  year      = {2015}
}

@article{mustafa2023active,
    author  = {Mustafa, Ahmad and AlRegib, Ghassan},
    title   = {Active Learning with Deep Autoencoders for Seismic Facies Interpretation},
    journal = {Geophysics},
    volume  = {88},
    number  = {4},
    pages   = {IM77--IM86},
    year    = {2023}
}

@article{wu2023mtlfaultnet,
    author  = {Wu, Weihua and Yang, Yang and Wu, Bangyu and Ma, Debo and Tang, Zhanxin and Yin, Xia},
    title   = {{MTL-FaultNet}: Seismic Data Reconstruction Assisted Multitask Deep Learning {3-D} Fault Interpretation},
    journal = {IEEE Transactions on Geoscience and Remote Sensing},
    volume  = {61},
    pages   = {1--15},
    year    = {2023}
}

@article{10449677,
  author={Mustafa, Ahmad and Rastegar, Reza and Brown, Tim and Nunes, Gregory and Delilla, Daniel and Alregib, Ghassan},
  journal={IEEE Transactions on Geoscience and Remote Sensing}, 
  title={Visual Attention-Guided Learning With Incomplete Labels for Seismic Fault Interpretation}, 
  year={2024},
  volume={62},
  number={},
  pages={1-12},
  doi={10.1109/TGRS.2024.3370037}}

\end{document}